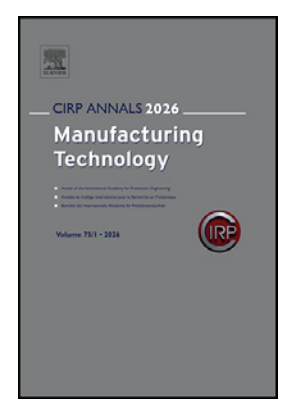

# AI-based worker guidance in assembly and disassembly operations using multimodal ego/exo-centric data capture and structured task knowledge

Vivek Chavan[a,b], Jörg Krüger (1)[a,b,*]

[a] *Automation Technology Division, Fraunhofer IPK, Berlin, Germany*
[b] *Department of Industrial Automation Technology, Technical University of Berlin, Germany*

ABSTRACT

Assembly and disassembly processes rely on expert knowledge that is difficult to document, reuse, and transfer. This paper presents a data-centric approach for extracting structured task knowledge from expert demonstrations using egocentric and exocentric recordings. Temporal and multimodal information from video and narration is jointly encoded to derive structured task representations that enable procedural documentation and context-aware worker guidance. The approach is evaluated on a real-world disassembly case study, demonstrating that video-based representations capture procedural structure and execution context beyond static image-based methods. The results highlight the potential of egocentric video understanding for repair, training, and circular manufacturing applications. Project website: https://indego-assistant.github.io/.

## 1. Introduction

Industrial assembly and disassembly operations face increasing challenges from high-mix, low-volume production and Circular Economy demands. Traditional automation struggles with contact-rich tasks involving tight tolerances, non-linear dynamics, and deformable objects (e.g., cables and wiring harnesses) [1,2]. These challenges are further amplified in disassembly, where end-of-life products often exhibit wear, damage, or missing components, making predefined CAD-based trajectories unreliable [3]. The absence of robust sensing, limited adaptability of rigid automation systems, and the high cost of sim-to-real transfer remain key obstacles to scalable and flexible automation in unstructured industrial environments [3,4].

Recent advances in Artificial Intelligence (AI) have led to the development of robust computer vision, machine learning, and natural language processing models capable of learning from large-scale data [1,5]. Foundation models, vision-language models, and world models have demonstrated strong generalization, in-context learning, and multimodal reasoning capabilities [5,6]. These models enable the joint processing of visual, linguistic, and temporal information and support higher-level reasoning over sequences of actions [6]. Generative and reasoning-based models allow inference and planning over structured representations, extending AI capabilities beyond perception toward procedural understanding and decision-making [5,6].

In production engineering and related domains, AI-based methods have been applied to perception, inspection, and decision support tasks, with increasing focus on human–robot collaboration and worker assistance [1,7]. Vision-language and large language models (VLMs/LLMs) have been used for scheduling, coordination, and monitoring in human-robot collaborative assembly, as well as for interpreting assembly steps and providing real-time guidance [8–10]. These approaches demonstrate the potential of multimodal AI to support complex industrial workflows, but their application is often limited to short-horizon tasks or static observations [11]. Long-horizon procedural tasks remain difficult to model due to their temporal complexity and context dependence [12].

Despite this progress, complex procedural tasks in assembly and disassembly still rely heavily on skilled human experts, contributing to productivity constraints and skilled worker shortages [12]. Existing systems lack effective mechanisms for capturing expert knowledge, extracting procedural structure, and reusing this knowledge for scalable worker support [13]. We address this gap by capturing rich multimodal data from expert demonstrations using combined egocentric (first-person) and exocentric (third person) data (Fig. 1). We define structured task knowledge with a precedence graph (PG) where nodes represent discrete actions and edges encode dependencies. Our approach operates on the assumption that expert narration serves as the primary ground truth for procedural intent. The framework enables context-aware worker guidance and next-step recommendation based on observed execution history. We validate the approach on a real-world case study on the disassembly of waste electrical and electronic equipment (WEEE) [14].

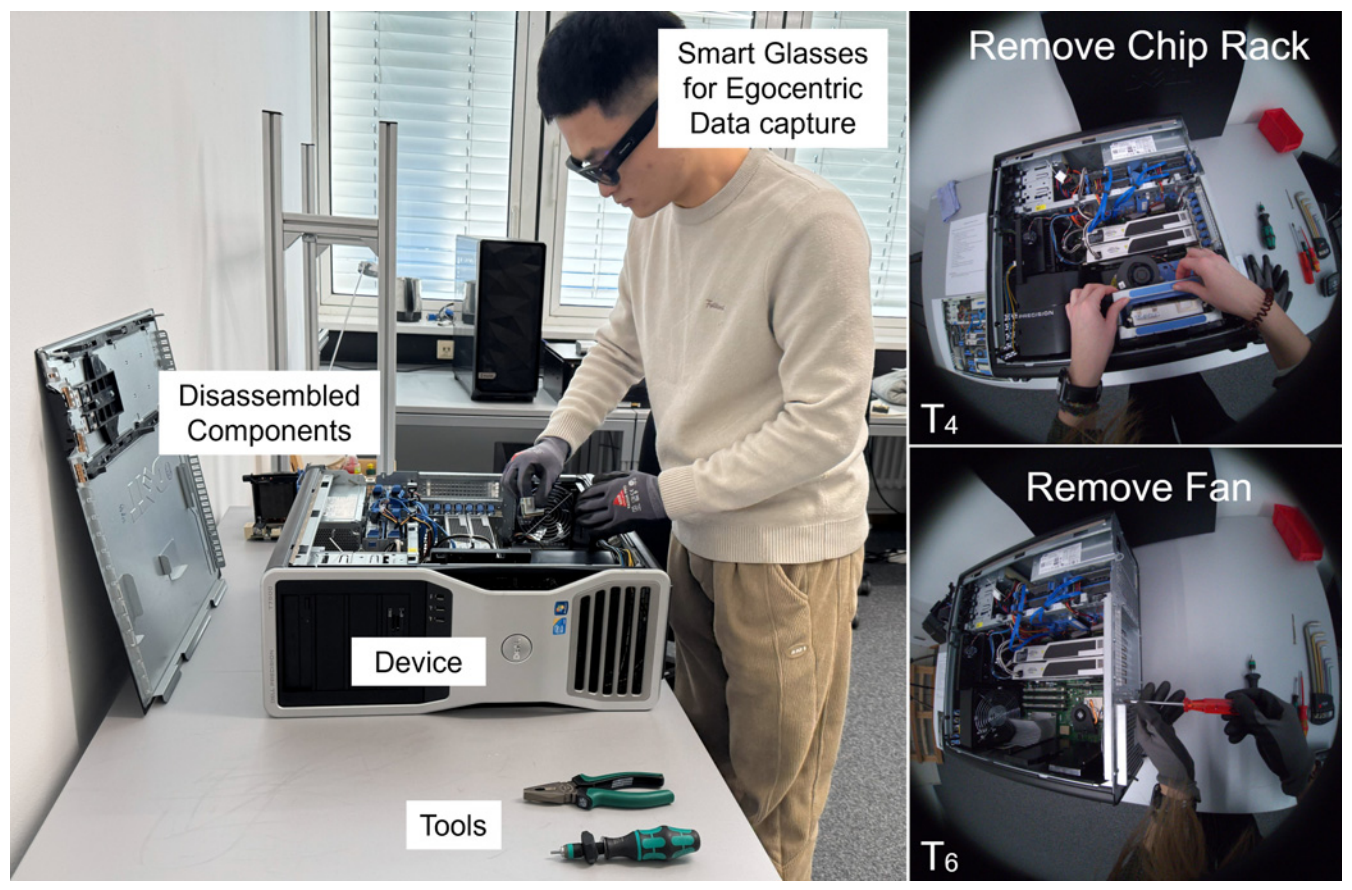

**Fig. 1.** (Left) Third-person exocentric view. (Right) First-person egocentric frames from a disassembly demonstration with corresponding procedural labels.

* Corresponding author.
*E-mail address:* joerg.krueger@tu-berlin.de (J. Krüger).

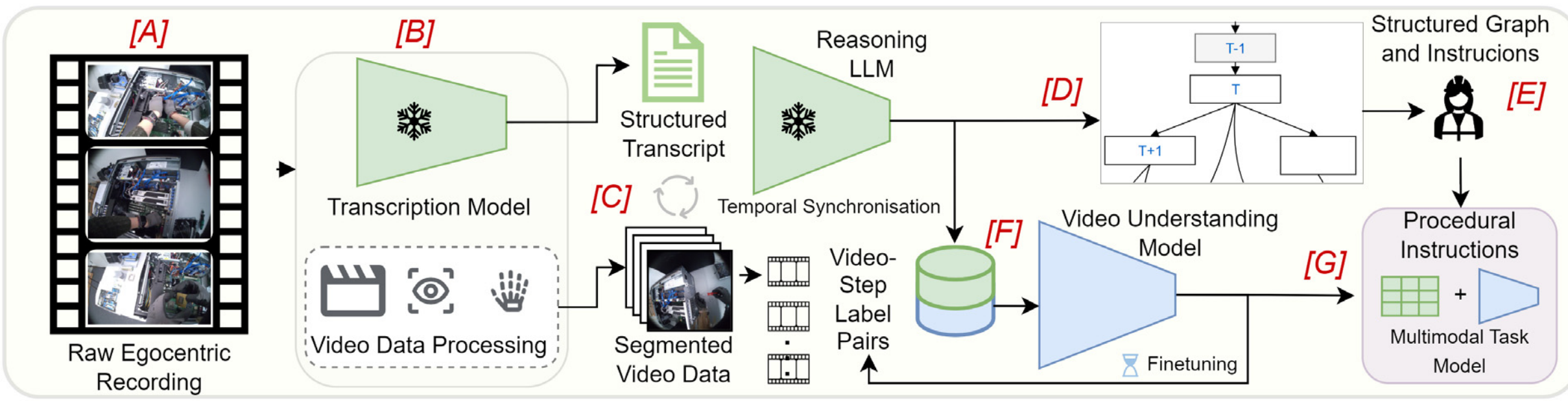

**Fig. 2.** System overview illustrating the transformation of expert egocentric raw recordings into structured task knowledge and trainable video representations ❄ indicates frozen models. ▪ corresponds to speech and natural language processing, ▪ corresponds to computer vision processing.

## 2. Methodology

This section describes the pipeline (Fig. 2) for capturing multimodal egocentric and exocentric data, extracting structured procedural knowledge, and enabling context-aware guidance.

### 2.1. Egocentric and exocentric data capture [A]

Egocentric sensing captures first-person visual and multimodal data from the user's perspective, enabling direct observation of hand–object interactions, tool usage, and action execution. It provides an unobstructed view and enables direct capture of complementary metadata such as motion signals, eye gaze, and narration [15]. Exocentric sensing provides a third-person view of the workspace, offering complementary scene-level context and spatial relationships [12]. Recent advances in computer vision and multimodal learning have increased interest in both paradigms for activity understanding, with relevance for industrial tasks that involve complex manual procedures [6,12,16,17].

We collect multimodal data using a combined egocentric–exocentric setup during expert task execution. Egocentric data is recorded using a smart glasses–based system [15] worn by the expert, capturing first-person video, eye gaze, and audio narration, as well as head and hand motion signals. Exocentric data is recorded using a stationary camera positioned to provide a stable overview of the workspace. The multimodal data collection setup is illustrated in Fig. 1. The exocentric view (left) captures the global scene context, including the operator, tools, and components. This is complemented by egocentric recordings (right) that provide a high-resolution view of hands-on interactions over time. The temporal labels shown in the egocentric frames correspond to the specific steps defined in the PG in Fig. 5.

Experts execute the procedure step by step while verbally describing actions, reasoning, constraints, and common errors. All sensor streams are recorded in a synchronized raw format and used as input for subsequent processing. This setup enables efficient capture of task-relevant multimodal data required for procedural modelling and worker guidance.

### 2.2. From ego/exo-centric recordings to structured task representation [B-E]

We define a task as a discrete procedural operation or keystep, and task structure as the logical dependencies connecting them. As illustrated in Fig. 2, expert narration from raw egocentric recordings is transcribed using WhisperX [18] (a foundation model trained with vast multilingual corpora of ego/exo data). This produces a timestamped transcript that temporally grounds the executed actions, explicitly linking the continuous vision input with the linguistic procedural intent and formal task structure.

The transcript is processed by LLaMA 3.1 (8B) [19], which leverages in-context learning and zero-shot reasoning to extract tasks and task structure. While an expert demonstration is serialized, the model derives non-linear dependencies and parallel branches by analyzing the narrated intent (e.g., "X must precede Y, but Z is independent"). This allows the automatic construction of a non-linear PG from a single sequential demonstration. Narration and video are then synchronized, segmenting the recording into step-level clips [C] assigned to semantic action labels.

The extracted procedural structure is represented as a PG modelled as a directed acyclic graph, where nodes denote procedural steps and edges encode precedence and dependency relations. The graph is reviewed and, if required, corrected by the expert to ensure procedural validity. After validation, the PG is converted into explicit procedural instructions and used as a structural prior for subsequent reasoning and guidance. The validated PG, together with the fine-tuned video understanding model [6], forms the basis for downstream action recognition and worker guidance.

### 2.3. Video understanding and reasoning [F, G]

The generated video-label pairs are used to obtain video representations for procedural action recognition. We employ the VJEPA-2 model [6], which is designed for video representation learning and demonstrates strong generalization across egocentric and exocentric data. When only a single expert recording is available, video embeddings are extracted from the pretrained backbone and used directly without fine-tuning. When multiple expert recordings are available, the model is fine-tuned by optimizing task-specific prediction heads on top of the extracted embeddings. These heads consist of either a lightweight (low-parameter) fully connected layer or a transformer-based encoder-decoder module. The segmented clips [C] serve as discrete units for the video understanding model [F], ensuring action recognition is grounded in validated temporal boundaries. The resulting model provides robust action recognition from video observations and serves as the perceptual input for decision making and worker guidance.

### 2.4. Worker guidance and task execution support [G-K]

During task execution, the system operates in an interactive closed-loop guidance mode, as shown in Fig. 3. Guidance can be triggered either explicitly by the worker or implicitly when task execution begins. Egocentric video is processed online and passed to the video understanding model to recognize procedural actions. Detected actions are accumulated into an execution history representing the steps completed so far. A reasoning-capable language model maps the execution history to the corresponding nodes of the structured PG thereby inferring the current procedural state. This allows the system to determine completed steps and valid next actions. Upon a guidance request, the reasoning component evaluates the PG constraints and generates an instruction for the next permissible step. The execution history is then updated, enabling continuous and context-aware guidance throughout task execution. Guidance is provided in text form and can be delivered through a graphical user interface, optionally accompanied by natural language narration. In addition, the interface may present retrieved example recordings from expert demonstrations illustrating the correct procedural step, supporting visual clarification of the recommended action. For tasks where the PG is already available, the video understanding component can still provide interactive feedback by grounding observed actions in the existing structure.

The framework can be extended with manufacturer documentation, structural priors, and component-level recognition modules linked to a digital product passport (DPP), requiring minimal additional processing and enabling applications in traceability and lifecycle-aware disassembly.

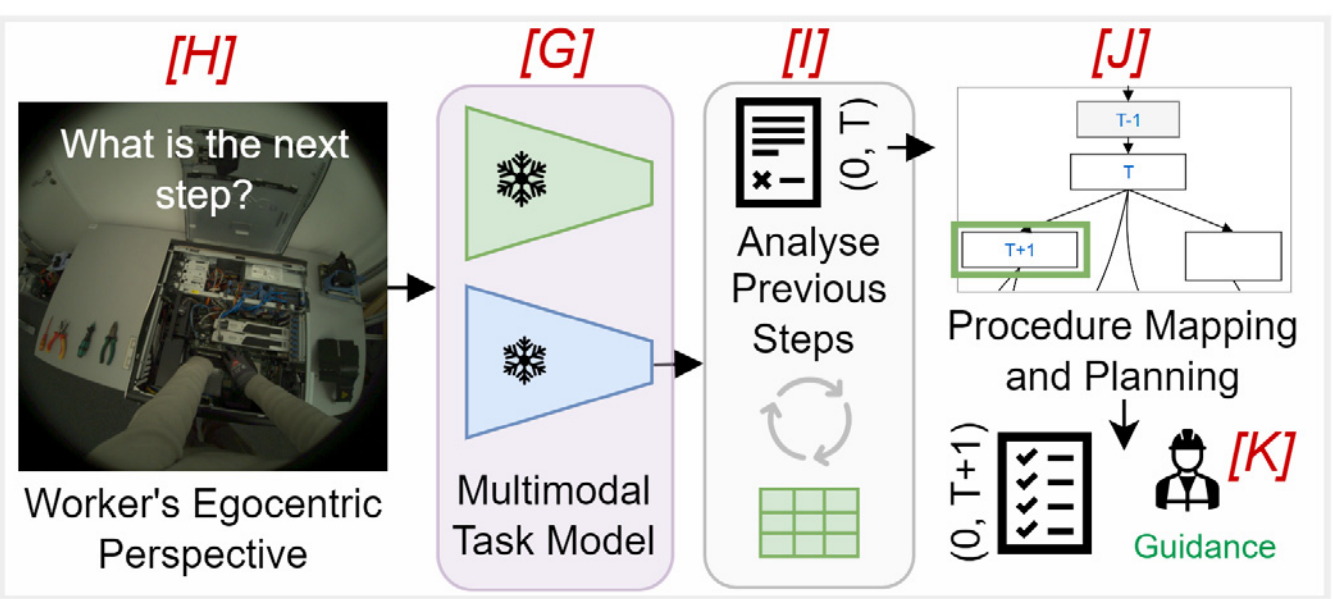

**Fig. 3.** Worker guidance phase in which the multimodal task model reasons over execution history and a structured PG to infer the next valid action and provide adaptive user guidance.

Fig. 4 outlines the computational logic for the end-to-end pipeline. During offline learning, raw expert recordings [A] are transcribed [B] and processed by a reasoning-based language model to extract procedural logic, temporal boundaries, and an initial PG [D]. These boundaries are used to segment the video [C], which, following expert validation of the PG [E], are used to adapt the video understanding model [F] to form the final multimodal task model [G]. During online execution, the system continuously monitors the worker's egocentric video stream [H]. Recognized actions update the execution history [I], enabling the language model to localize the current state within the PG [J] and output valid next-step instructions [K].

Algorithm: End-to-End Procedure Learning and Worker Guidance

```
#Models: WhisperX (Speech), LLM (Language), v_model
 (Video)
#Inputs: raw_audio, raw_video from expert demonstrations

#Phase 1: Procedure Learning from Expert Data [A-G]
raw_audio, raw_video = capture_expert_demo() # [A]
transcript = WhisperX.transcribe(raw_audio) # [B]
labels, time_bounds, initial_PG
 =LLM.extract_logic(transcript) # [D]

labeled_segments = segment_video(raw_video, time_bounds,
 labels) # [C]
valid_PG = expert_review(initial_PG) # [E]

# Adapt video model based on available data
if num_demos > 1:
    V_model.finetune(labeled_segments) # [F]
else:
    memory_bank =
 V_model.extract_embeddings(labeled_segments) # [F]
task_model = finalize_package(valid_PG, V_model) # [G]

#Phase 2: Context-Aware Worker Guidance [H-K]
history = [] # [I] Setup history

for live_clip in worker_video_stream: # [H] Live capture
    # Action recognition (prediction or retrieval)
    if num_demos > 1:
        current_step = V_model.predict(live_clip)
    else:
        current_step =
 nearest_neighbor(V_model(live_clip), memory_bank)

    # Update execution state and provide guidance
    if current_step and current_step not in history:
        history.append(current_step) # [I] Update state
    next_valid_steps = LLM.infer_next(valid_PG, history)
 # [J] Plan next
    display_instructions(next_valid_steps) # [K] Output
```

**Fig. 4.** Python-style pseudocode implementation of the framework. Annotations [A-K] map directly to the stages in Fig. 2.

## 3. Case study

This section presents a real-world case study used to evaluate the proposed approach, including the application scenario, dataset, and experimental setup for WEEE disassembly. Disassembly poses challenges that differ fundamentally from assembly. While assembly typically follows standardized and predefined sequences, disassembly must address uncertainty arising from product wear, missing components, undocumented modifications, and diverse end-of-life conditions and variants [20]. This variability limits the applicability of rigid, product-specific procedures and complicates process modelling.

### 3.1. IndEgo dataset

The proposed approach is evaluated on a real-world disassembly scenario from the IndEgo dataset [12], which comprises multimodal egocentric and exocentric recordings of industrial assembly and disassembly tasks performed by expert users in laboratory environments [21]. Each recording includes synchronized first-person video, third-person video, and spoken expert narration describing executed actions and procedural reasoning.

For this case study, we focus on the disassembly of a workstation computer as a representative WEEE processing task. The selected scenario involves multiple interacting components, tool usage, and nontrivial procedural dependencies, making it suitable for evaluating structured task representation, video understanding, and worker guidance. Expert recordings are used to derive video-label pairs, PGs, and instructions. For verification and validation, the AI-generated outputs are compared against manually annotated ground-truth sequences to ensure procedural accuracy.

### 3.2. Workstation disassembly task

The disassembly procedure (illustrated in Fig. 1) is formalized as the structured PG shown in Fig. 5. This representation is derived from the automated pipeline described in Section 2, specifically following the human-in-the-loop validation step [E], where an expert reviews and corrects the generated graph to ensure procedural accuracy. The final PG encodes the execution flow from preparation to cleanup, capturing both strict sequential dependencies (e.g., Cabinet → Duct → RAM) and interchangeable component-removal steps via parallel branches. This structure reflects real-world procedural flexibility while providing a formal foundation for systematic execution monitoring and safety-critical verification.

The procedure starts with a preparation phase, including tool collection, safety compliance, and verification of suitable workplace conditions. The main disassembly phase involves removal of internal components such as storage devices, cooling elements, processing units, and auxiliary hardware. Several steps may be executed in varying order, while others require fixed precedence due to hardware constraints. The procedure concludes with cleanup and verification, including completion checks, component sorting, and tool return. Figs. 1 and 5 show representative egocentric frames corresponding to selected PG steps. The first-person views illustrate variability in hand–object interactions, viewpoints, and component visibility across the procedure. This variability highlights the procedural complexity of the disassembly task and motivates the use of structured task representations and context-aware video understanding for reliable execution monitoring and worker guidance.

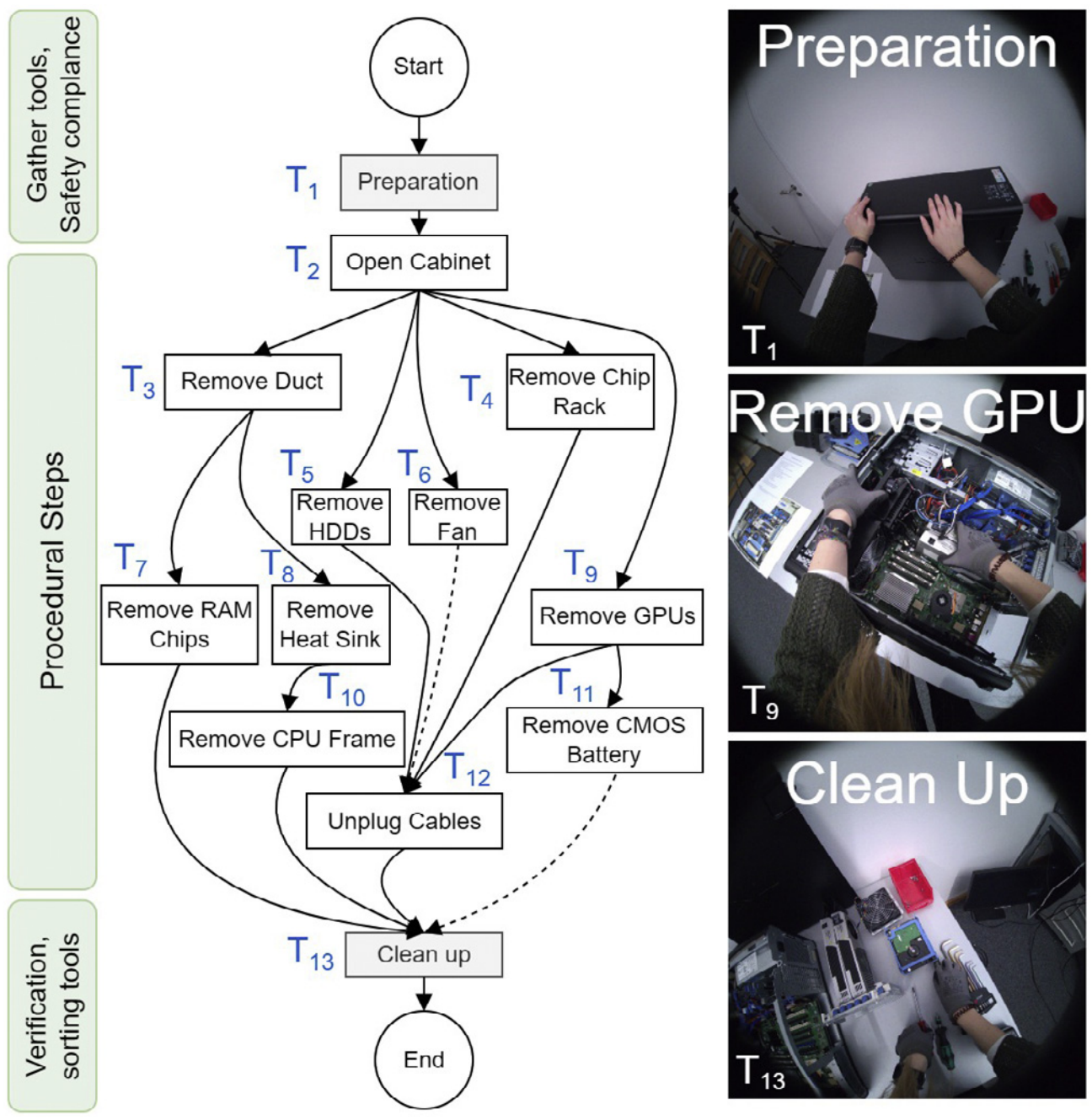

**Fig. 5.** Corrected Precedence Graph (PG) for the workstation disassembly task, illustrating sequential precedence constraints and parallel component-removal branches (— shows edges not in initial PG).

## 4. Experimental evaluation

The proposed pipeline is evaluated on a real-world workstation disassembly task using egocentric recordings from the IndEgo dataset. All processing is performed offline on a dedicated workstation equipped with an NVIDIA A6000 GPU. Fine-tuning the video understanding model requires approximately 12 GB of GPU memory, while the language-model-based reasoning component requires approximately 5 GB without quantization. Each expert recording has a duration of approximately 20 min.

Two experimental configurations are considered. In the first, a single expert demonstration is available, and video embeddings are extracted without fine-tuning. In the second, four expert recordings captured under slightly varying conditions are used to fine-tune the video model. In both settings, recordings are processed to generate structured transcripts, video–label pairs, PG, and procedural instructions. Worker guidance is evaluated on a separate execution recording by sequentially mapping observed actions to the PG to infer the current procedural state and generate next-step recommendations.

The results, summarized in Table 1, show high intersection-over-union (IoU) scores for video–label pair generation, indicating strong agreement with manual annotations. PG generation achieves high step coverage and edge-level F1 scores, demonstrating reliable extraction of procedural steps and dependencies. Even with a single expert demonstration, the system derives meaningful procedural structure that supports action recognition and worker guidance, while multiple demonstrations improve recognition accuracy through increased robustness and variability coverage. From a usability perspective, the achieved next-step guidance accuracy is sufficient for reliable assistive recommendations, with occasional recognition errors tolerated without disrupting overall procedural flow.

End-to-end processing of a complete expert recording requires approximately 1 h to produce the multimodal task model. During deployment, real-time performance depends on system latency and the user interface, which are outside the scope of this work. Nevertheless, the results indicate that structured PG-based reasoning enables effective execution monitoring and next-step guidance under realistic conditions. This performance level is suitable for advisory worker assistance, where guidance complements human decision-making rather than enforcing fully autonomous control, supporting practical adoption in industrial (dis-)assembly and other procedural tasks.

**Table 1**
Quantitative evaluation of the main stages of the proposed pipeline.

| Objective | Metric | Result | |
|---|---|---|---|
| | | 1 Demo | 4 Demos |
| Video–label pairs | IoU overlap | 0.84 | 0.92 |
| Precedence graph generation | Edge F1 | 0.79 | 0.87 |
| | Step coverage | 0.88 | 0.93 |
| Video model prediction | Validation Accuracy | – | 0.81 |
| | Action recognition accuracy | 0.66 | 0.80 |
| Worker guidance | Next-step Accuracy | 0.76 | 0.84 |

A higher score is better.

## 5. . Conclusions and outlook

This paper demonstrates how recent advances in egocentric data capture, multimodal video understanding, and reasoning-based task modelling enable data-driven generation of structured assistance for assembly and disassembly processes. The proposed approach automatically extracts procedural knowledge from expert demonstrations and supports context-aware worker guidance during task execution. Validation on a real-world workstation disassembly case study shows that the system can reliably recognize actions, construct PGs, and recommend valid next steps. The results indicate that meaningful procedural guidance can be derived even from a single expert demonstration, with improved robustness when multiple recordings are available. The approach is particularly valuable for disassembly and reassembly scenarios, where explicit procedural knowledge such as PG is typically unavailable.

The presented method highlights the growing potential of AI-driven techniques to efficiently generate complex assistance and automation functions for (dis-)assembly processes. Due to its high degree of automation, the effort required to create a procedural guidance model is comparable to performing a single example procedure, with only a small increase in data requirements for complex perspectives or tasks. Future work will explore extensions towards real-time mistake detection, generalized procedural understanding across classes of similar but non-identical products, and the transfer of egocentric procedural knowledge to robotic learning for (dis-)assembly, enabling long-horizon planning and automation based on human demonstrations [22,23].

## Declaration of competing interest

The authors declare that they have no known competing financial interests or personal relationships that could have appeared to influence the work reported in this paper.

## CRediT authorship contribution statement

**Vivek Chavan:** Writing – review & editing, Writing – original draft, Visualization, Validation, Software, Methodology, Investigation, Data curation, Conceptualization. **Jörg Krüger (1):** Writing – review & editing, Writing – original draft, Validation, Supervision, Resources, Methodology, Funding acquisition, Formal analysis, Conceptualization.

## Acknowledgements

This work is partially funded by the German Federal Ministry of Research, Technology and Space (BMFTR) and the German Aerospace Center (DLR) under the KIKERP project (Grant No 16IS23055C) in the KI4KMU program. This work was also supported by the Fraunhofer Internal Programs under Grant No. SME 40-12767.

## References

[1] Gao RX, Krüger J, Merklein M, Möhring H-C, Vancza J (2024) Artificial Intelligence in Manufacturing: State of the Art, Perspectives, and Future Directions. *CIRP Annals* 73(2):723–749.

[2] Tang C, Abbatematteo B, Hu J, Chandra R, Martín-Martín R, Stone P (2025) Deep Reinforcement Learning for Robotics: A Survey of Real-World Successes. *Annual Review of Control, Robotics, and Autonomous Systems* 8:153–188.

[3] Kaspar M, Muñoz Osorio JD, Bock J (2020) Sim2real Transfer for Reinforcement Learning Without Dynamics Randomization. In: *Proceedings of the 2020 IEEE/RSJ International Conference on Intelligent Robots and Systems (IROS)*, Piscataway, NJ. IEEE, , 4383–4388.

[4] Tobin J, Fong R, Ray A, Schneider J, Zaremba W, Abbeel P (2017) Domain Randomization for Transferring Deep Neural Networks from Simulation to the Real World. In: *Proceedings of the 2017 IEEE/RSJ International Conference on Intelligent Robots and Systems (IROS)*, Vancouver, BC. IEEE, , 23–30.

[5] Brown TB, Mann B, Ryder N, Subbiah M, Kaplan J, Dhariwal P, et al. (2020) Language Models are Few-Shot Learners. *Advances in Neural Information Processing Systems* 33:1877–1901.

[6] Assran M., Bardes A., Fan D., Garrido Q., Howes R., Komeili M., et al. (2025) V-JEPA 2: Self-Supervised Video Models Enable Understanding, Prediction and Planning. arXiv preprint arXiv:2506.09985.

[7] Gkournelos C, Konstantinou C, Makris S (2024) An LLM-Based Approach for Enabling Seamless Human-Robot Collaboration in Assembly. *CIRP Annals* 73 (1):9–12.

[8] Yin Y, Wan K, Li C, Zheng P (2025) An LLM-Enabled Human Demonstration-Assisted Hybrid Robot Skill Synthesis Approach for Human-Robot Collaborative Assembly. *CIRP Annals* 74(1):1–5.

[9] Dimitropoulos N, Kaipis M, Giartzas S, Michalos G (2025) Generative AI for Automated Task Modelling and Task Allocation in Human Robot Collaborative Applications. *CIRP Annals* 74(1):7–11.

[10] Simeone A, Fan Y, Antonelli D, Priarone PC, Settineri L (2025) Conceptualisation of a Multimodal, Non-Intrusive, Generative AI-Based Assistive System for Assembly. *CIRP Annals* 74(1):37–41.

[11] Urgo M, Tarabini M, Tolio T (2019) A Human Modelling and Monitoring Approach to Support the Execution of Manufacturing Operations. *CIRP Annals* 68(1):5–8.

[12] Chavan V, Imgrund Y, Dao T, Bai S, Wang B, Lu Z, Heimann O, Krüger J (2025) IndEgo: A Dataset of Industrial Scenarios and Collaborative Work for Egocentric Assistants. *Advances in Neural Information Processing Systems* 38.

[13] Addepalli S, Weyde T, Namoano B, Oyedeji OA, Wang T, Erkoyuncu JA, Roy R (2023) Automation of Knowledge Extraction for Degradation Analysis. *CIRP Annals* 72(1):33–36.

[14] Duflou JR, Peeters JR, Altamirano D, Bracquene E, Dewulf W (2018) Demanufacturing Photovoltaic Panels: Comparison of End-of-Life Treatment Strategies for Improved Resource Recovery. *CIRP Annals* 67(1):29–32.

[15] Engel J., Somasundaram K., Goesele M., Sun A., Gamino A., Turner A., et al. (2023) Project Aria: A New Tool for Egocentric Multi-Modal AI Research. arXiv preprint arXiv:2308.13561.

[16] Sener F, Chatterjee D, Shelepov D, He K, Singhania D, Wang R, Yao A (2022) Assembly101: A Large-Scale Multi-View Video Dataset for Understanding Procedural Activities. In: *Proceedings of the IEEE/CVF Conference on Computer Vision and Pattern Recognition (CVPR)*, IEEE, New Orleans, LA, USA, 10334–10345.

[17] Plizzari C, Goletto G, Furnari A, Bansal S, Ragusa F, Farinella GM, Damen D, Tommasi T (2024) An Outlook Into the Future of Egocentric Vision. *International Journal of Computer Vision* 132(11):4880–4936.

[18] Bain M, Huh J, Han T, Zisserman A (2023) WhisperX: Time-Accurate Speech Transcription of Long-Form Audio. *INTERSPEECH* : 1–5.

[19] Grattafiori A., Dubey A., Jauhri A., Pandey A., Kadian A., Al-Dahle A., et al. (2024) The Llama 3 Herd of Models. arXiv preprint arXiv:2407.21783.

[20] Vongbunyong S, Kara S, Pagnucco M (2013) Application of Cognitive Robotics in Disassembly of Products. *CIRP Annals* 62(1):31–34.

[21] Chavan V, Heimann O, Krüger J (2025) On the Application of Egocentric Computer Vision to Industrial Inspection. *ECCV 2024 Workshops. Lecture Notes in Computer Science*, Springer, Cham, 1–17.

[22] Figure A.I. (2024) Project Go-big: Building Internet-Scale Data for Humanoid Robots. Figure AI News. Accessed December 19, 2025 At: https://www.figure.ai/news/project-go-big.

[23] Sermanet P., Lynch C., Hsu J., Levine S. (2017) Time-Contrastive Networks: Self-Supervised Learning from Multi-View Observation. CVPR Workshops:486–487.